\documentclass{article}

\usepackage{arxiv}

\usepackage[utf8]{inputenc} 
\usepackage[T1]{fontenc}    
\usepackage{hyperref}       
\usepackage{url}            
\usepackage{booktabs}       
\usepackage{amsfonts}       
\usepackage{nicefrac}       
\usepackage{microtype}      
\usepackage{lipsum}		
\usepackage{caption} 
\usepackage{graphicx}
\usepackage[table,xcdraw]{xcolor} 
\usepackage{doi}

\usepackage{amsmath}
\usepackage{listings}
\usepackage{xcolor} 
\definecolor{keywordcolor}{rgb}{0.0, 0.0, 1.0} 
\definecolor{commentcolor}{rgb}{0.0, 0.5, 0.0} 
\definecolor{stringcolor}{rgb}{1.0, 0.0, 0.0} 
\title{Evaluating Tokenizer Performance of Large Language Models Across Official Indian Languages}

\author{ \href{https://orcid.org/0009-0007-8038-1278}{\includegraphics[scale=0.06]{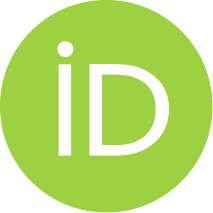}\hspace{1mm}Sagar Tamang}\thanks{
    Correspondance can be addressed to \textit{cs22bcagn033@kazirangauniversity.in}} \\
	Department of IT\\
	The Assam Kaziranga University\\
	Jorhat, India \\
	\texttt{cs22bcagn033@kazirangauniversity.in} \\
	\And
	\href{https://orcid.org/0000-0001-7809-5220}{\includegraphics[scale=0.06]{orcid.pdf}\hspace{1mm}Dr. Dibya Jyoti Bora} \\
	Department of IT\\
	The Assam Kaziranga University\\
	Jorhat, India \\
	\texttt{dibyajyotibora@kazirangauniversity.in} \\
}

\renewcommand{\shorttitle}{Evaluating Tokenizer Performance of Large Language Models in Indian Languages}

\hypersetup{
pdftitle={\textbf{Evaluating Tokenizer Performance of Large Language Models Across Official Indian Languages}},
pdfsubject={cs.AI},
pdfauthor={S. Tamang and D. J. Bora},
pdfkeywords={tokenizer, LLM, tokens, GPT, Assamese, SUTRA},
}

\begin{document}
\maketitle

\begin{abstract}
Large Language Models (LLMs) based on transformer architectures have revolutionized a variety of domains, with tokenization playing a pivotal role in their pre-processing and fine-tuning stages. In multilingual models, particularly those tailored for Indic languages, effective tokenization is crucial for optimizing performance. This paper presents a comprehensive evaluation of tokenizers used by 12 LLMs across all 22 official languages of India, with a focus on comparing the efficiency of their tokenization processes. We employed the Normalized Sequence Length (NSL) as a key metric in our analysis. Our findings reveal that the SUTRA tokenizer outperforms all other models, including several Indic-specific models, excelling in 14 languages. Notable insights include the SUTRA tokenizer's superior handling of Indic languages, GPT-4o's advancement over its predecessor GPT-4 in processing Indian languages, and the limited performance of Project Indus in certain languages. This study underscores the critical importance of developing targeted tokenization strategies for multilingual and Indic-centric models, laying the groundwork for future improvements in tokenizer design to enhance linguistic coverage and model efficiency.
\end{abstract}
\section{Introduction}
\label{sec:introduction}
\subsection{Background}
In an ever-evolving landscape of Artificial Intelligence (AI), transformers-based generative Large Language Models (LLMs) are transforming an increasing number of fields with an ever-increasing number of applications in finance, medicine, education, and many more \cite{CHIARELLO2024103002,nie2024surveylargelanguagemodels}. Tokenization is an important step for LLMs, especially in pre-processing and fine-tuning stages \cite{tamang2024performanceevaluationtokenizerslarge}. 

\begin{figure}[h!]
    \centering
    \includegraphics[width=0.8\textwidth]{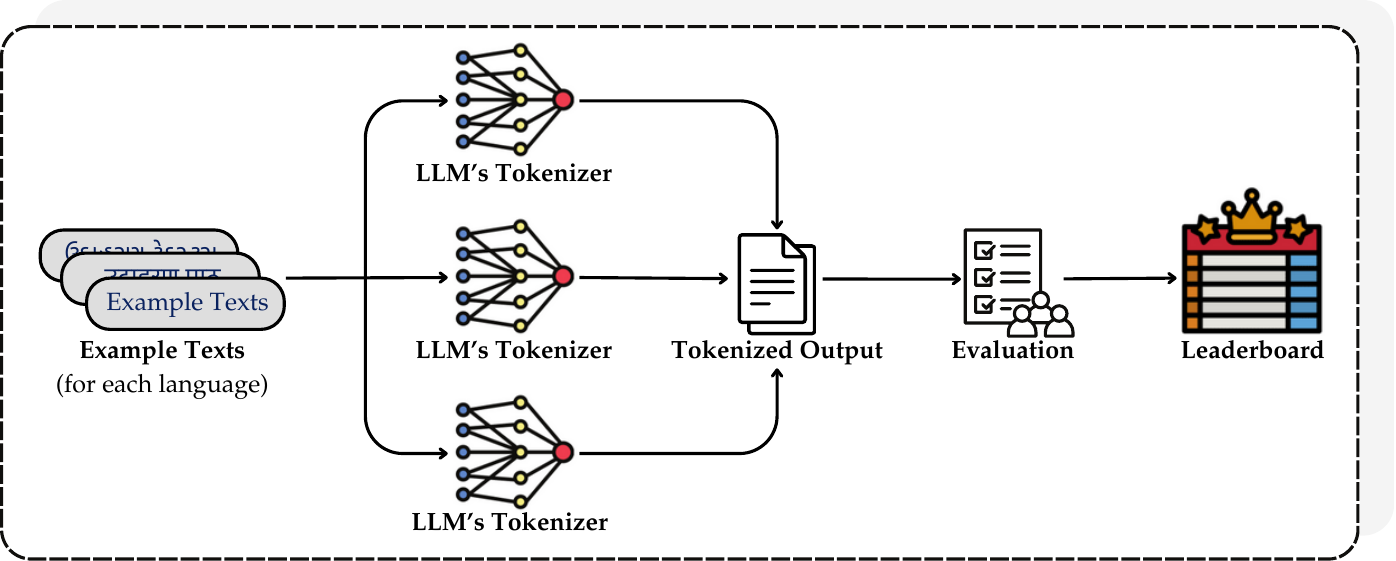}
    \caption{Evaluation pipeline: (1) We collect example texts for all 22 languages. (2) We send the example texts to the LLMs' tokenizer. (3) Evaluate the tokenized outputs. (4) We construct leaderboards using our evaluation.}
    \label{fig:evaluation-step}
\end{figure}

Most of the LLMs use either of two types of tokenization algorithms, namely WordPiece and Byte Pair Encoding (BPE). For example, OpenaAi's GPT-4o model and META's Llama 3, both employ a modified BPE tokenizer \cite{yang2024largelanguagemodeltokenizer}. WordPiece was developed for models like BERT employing a greedy approach. It starts with the longest substring that matches a token in its vocabulary, allowing it to handle out-of-vocabulary words effectively by breaking them down into known subword units \cite{song2021fastwordpiecetokenization,ogundepo2022betterwhitespaceinformationretrieval}.  BPE works by iteratively merging the most frequently occurring pairs of characters or subwords in a corpus to create a complete vocabulary \cite{kozma2024theoreticalanalysisbytepairencoding,zouhar2024formalperspectivebytepairencoding}.

Agnostic of the tokenization algorithms used, many techniques have been developed to compare tokenizers of LLMs. Subword fertility is one such technique which measures the average number of tokens used per word \cite{singh2024indicgenbenchmultilingualbenchmarkevaluate}. Normalized Sequence Length (NSL) is another such metric used to evaluate the efficiency of tokenizers \cite{dagan2024gettingtokenizerpretrainingdomain}. 

An effective tokenizer is essential for enabling a model to learn a language efficiently. Improved tokenizer performance offers several benefits, including faster token generation and reduced computational resource requirements, enhancing both efficiency and cost-effectiveness \cite{microsoft2024gpt4o}.

\subsection{Tokenizers in Multilingual \& Indic Models}
Many multilingual models, including OpenAI's ChatGPT \cite{openai2024gpt4technicalreport}, Google's Gemini \cite{geminiteam2024geminifamilyhighlycapable}, Meta's Llama \cite{dubey2024llama3herdmodels}, and TWO AI's SUTRA \cite{bendale2024sutrascalablemultilinguallanguage}, are designed to deliver coherent performance across a wide range of global languages, including Indian languages. Achieving this requires tokenizers in these LLMs to handle diverse languages efficiently. This study evaluates the performance of tokenizers in these multilingual models and Indic-specific models across all official languages of India.

The rest of the paper is organized in the following way: Section \ref{sec:literature} highlights related works, Section \ref{sec:methodology} describes the methodology of this study, Section \ref{sec:results} is where the results are showcased and Section \ref{sec:discussion} is for discussions with future outlook.
\section{Literature Review}
\label{sec:literature}

\subsection{Large Language Models and Tokenization}
Ever since the release of transformer-based architecture, Large Language Models (LLMs) have demonstrated remarkable capabilities in Natual Language Processing (NLP) tasks and beyond \cite{naveed2024comprehensiveoverviewlargelanguage}. However, some recent studies \cite{minaee2024largelanguagemodelssurvey} have also proposed promising non-transformer LLMs. LLMs today exhibit a plethora of applications not limited to NLP tasks, but can also perform general tasks performing multi-step reasoning. Thus, LLMs are becoming the basic building block for developing general-purpose AI agents or Artificial General Intelligence (AGI) \cite{minaee2024largelanguagemodelssurvey}. 

Tokenization refers to the process of converting a sequence of texts into smaller parts, known as tokens. In order to increase the coverage of dictionaries and also to deal with words that were unseen in training data, LLMs use sub-words based tokenizers like BPE or Wordpiece \cite{yang2024largelanguagemodeltokenizer,zouhar2024formalperspectivebytepairencoding,kozma2024theoreticalanalysisbytepairencoding,minaee2024largelanguagemodelssurvey}. 

\subsection{Indic-Specific Language Models}
S. Bhat et al. evaluated the capabilities of generative models like ChatGPT, mT0, and BLOOMZ in generating Indic languages, and the findings revealed that these models have limited capabilities in generating text in Indic languages in a zero-shot setting. While they performed better on manual quality evaluations for English, their performance in Indic languages highlighted the need for further development and training specifically tailored to these languages \cite{bhat2023indic}. Several studies have introduced benchmarks to evaluate the performance of LLMs across Indic languages, and their results emphasize the necessity for more focused research and development efforts in LLMs to handle the linguistic diversity of India \cite{singh2024indicgenbench,singh2024indicqabench,kumar2022indicnlg}. 

One notable advancement is SUTRA (Scalable Multilingual Language Model Architecture), introduced by A. Bendale et al. SUTRA supports over 50 languages, including Indic ones, by decoupling conceptual understanding from language-specific processing for scalable multilingual learning. It uses a Mixture of Experts framework, enhancing computational efficiency and responsiveness. Evaluations show that SUTRA outperforms models like GPT-3.5 and Llama2 by 20-30\% on MMLU benchmarks. The model provides hallucination-free, factual, and up-to-date responses, with the potential to democratize AI access, especially in non-English regions \cite{bendale2024sutrascalablemultilinguallanguage,twoai_sutra}.

\subsection{Comparative Evaluation of Tokenizers}
To compare the tokenizers, several metrics and methodologies have been introduced. Subword fertility is one such metric that measures the average number of tokens generated per word \cite{dagan2024gettingtokenizerpretrainingdomain}. An ideal tokenizer in this case would have a fertility of 1.0, indicating that most words are represented as single tokens, while a higher fertility score signals that many words are split into multiple tokens \cite{euTokenizerPerformance}. Another metric that is employed is the proportion of continued words \cite{euTokenizerPerformance} which indicates the percentage of words that are split into multiple tokens. 0 is an ideal value for the proportion of continued words. A lower proportion indicates better performance, as it means fewer words are fragmented \cite{euTokenizerPerformance}. Another metric that is used to evaluate and compare tokenizers is the Normalized Sequence Length (NSL). NSL measures the average length of tokenized sequences produced by a tokenizer relative to a baseline tokenizer \cite{dagan2024gettingtokenizerpretrainingdomain}.

Existing tokenizer studies are performed at a smaller scale than what is typical for modern LLMs or focus on multilingual tasks \cite{dagan2024gettingtokenizerpretrainingdomain}. Some studies have tried to evaluate the performance of tokenizers but only in some select languages like Turkish \cite{toraman2023impacttokenization}, Arabic \cite{alyafeai2021evaluatingtokenizers}, or focusing on a few Indian languages \cite{tamang2024performanceevaluationtokenizerslarge,adasci2024multilingualtokenization}. 

\subsection{Gaps in Existing Research}
Thus, we can observe that there exists a research gap to evaluate the tokenizers in other Indian languages as well. Hence, in this study, we are conducting an overall evaluation of the tokenizers of LLMs in all 22 official languages of India as recognized by the eighth schedule of the Indian constitution \cite{eighthSchedule}.
\section{Methodology}
\label{sec:methodology}
Our evaluation step is summarized in Figure \ref{fig:evaluation-step}.

\subsection{Example Texts}
\begin{figure}[h!]
    \centering
    \includegraphics[width=0.5\textwidth]{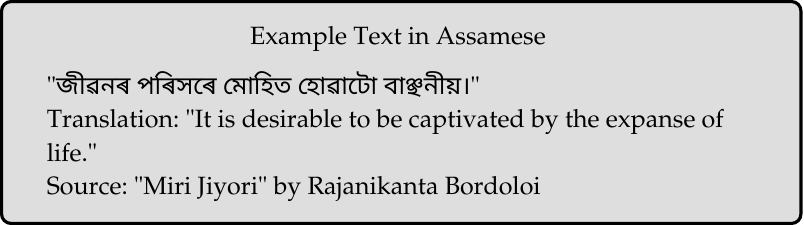}
    \caption{Assamese text used for evaluating tokenizer performance.}
    \label{fig:example-assamese}
\end{figure}

We compiled example texts in all 22 languages to evaluate the performance of the tokenizers. Each language's text was selected in its primary writing script to ensure an authentic assessment of the tokenizer's capability to process native scripts accurately. The curated example texts represent diverse linguistic structures and scripts, enabling a comprehensive analysis of tokenization performance.

All the example texts used during our study can be found in the Appendix \ref{sec:appendix-b} or Figure \ref{fig:example-texts-1st-half} and \ref{fig:example-texts-2nd-half}. One such example text can be found in Figure \ref{fig:example-assamese}.

\subsection{Models}
We chose 12 models including proprietary multilingual models, as well as Open-weights multilingual and Indic language models for our study. The list of models can be found in Table \ref{tab: tokenizer_list}.

While we acknowledge that some models are not specifically designed for all Indian languages—such as \textit{MahaMarathi}, which is tailored for Marathi, and \textit{MBZUAI's Nanda}, which is optimized for Hindi and English—we have included them in this study for the sake of comprehensive evaluation.

\begin{table*}[!ht]
\centering 
\small
\begin{tabular}{@{}lcc@{}}
\toprule
\textbf{Models} & \textbf{\begin{tabular}[c]{@{}c@{}}Languages \\ Tested\end{tabular}} & \textbf{Availability} \\ \midrule
\href{https://huggingface.co/Xenova/gpt-4o}{\textit{GPT-4o}} & All & Proprietary \\
\href{https://huggingface.co/Xenova/gpt-4}{\textit{GPT-4}} & All & Proprietary \\
\href{https://huggingface.co/TWO/sutra-mlt256-v2}{\textit{TWO/sutra-mlt256-v2}} & All & Proprietary \\
\href{https://huggingface.co/microsoft/Phi-3.5-MoE-instruct}{\textit{microsoft/Phi-3.5-MoE-instruct}} & All & Open-weights \\
\href{https://huggingface.co/meta-llama/Llama-3.1-405B-FP8}{\textit{meta-llama/Llama-3.1-405B-FP8}} & All & Open-weights \\
\href{https://huggingface.co/ai4bharat/Airavata}{\textit{ai4bharat/Airavata}} & All & Open-weights \\
\href{https://huggingface.co/CohereForAI/aya-23-35B}{\textit{CohereForAI/aya-23-35B}} & All & Open-weights \\
\href{https://huggingface.co/MBZUAI/Llama-3-Nanda-10B-Chat}{\textit{MBZUAI/Llama-3-Nanda-10B-Chat}} & All & Open-weights \\
\href{https://huggingface.co/nickmalhotra/ProjectIndus}{\textit{nickmalhotra/ProjectIndus}} & All & Open-weights \\
\href{https://huggingface.co/sarvamai/OpenHathi-7B-Hi-v0.1-Base}{\textit{sarvamai/OpenHathi-7B-Hi-v0.1-Base}} & All & Open-weights \\
\href{https://huggingface.co/Telugu-LLM-Labs/Indic-gemma-7b-finetuned-sft-Navarasa-2.0}{\textit{Telugu-LLM-Labs/Indic-gemma-7b-finetuned-sft-Navarasa-2.0}} & All & Open-weights \\
\href{https://huggingface.co/marathi-llm/MahaMarathi-7B-v24.01-Base}{\textit{marathi-llm/MahaMarathi-7B-v24.01-Base}} & All & Open-weights \\ \bottomrule
\end{tabular}
\captionsetup{belowskip=12pt} 
\caption{List of tokenizers tested. ``All" refers to all 22 official languages of India as recognized by the Eighth Schedule of the Indian Constitution. The official languages include Assamese, Bengali, Bodo, Dogri, Gujarati, Hindi, Kannada, Kashmiri, Konkani, Maithili, Malayalam, Manipuri, Marathi, Nepali, Odia, Punjabi, Sanskrit, Santali, Sindhi, Tamil, Telugu, Urdu.}
\label{tab: tokenizer_list}
\end{table*}

\subsection{Evaluation Metric}
For our work is extending the previous works by \cite{tamang2024performanceevaluationtokenizerslarge}, we have chosen to go with the NSL metric. Formally the NSL is defined by \cite{dagan2024gettingtokenizerpretrainingdomain} \( c_{\lambda \beta} \) as the ratio between the length of an encoded sequence from a tokenizer \( T_\lambda \) and a tokenizer \( T_\beta \). For \( N \) examples taken from a dataset \( D \):

\[
c_{\lambda \beta} = \frac{\sum_{i=1}^{N} \text{length}(T_\lambda(D_i))}{\sum_{i=1}^{N} \text{length}(T_\beta(D_i))}
\]

\section{Results}
\label{sec:results}
\textbf{Average NSL Values}
Table \ref{tab:avg-nsl-values} presents the average NSL values for all tokenizers across the 22 languages, calculated using the examples provided in Appendix \ref{sec:appendix-b}. The scores are reported to four decimal places for precision. The \textbf{bold text} indicates the lowest value or the best performance among all the other tokenizers. It can be observed that SUTRA tokenizer manages to excel among all the other tokenizers including ChatGPT's 4-o or other Indic models. 

\begin{figure}[h!]
    \centering
    \includegraphics[width=0.5\textwidth]{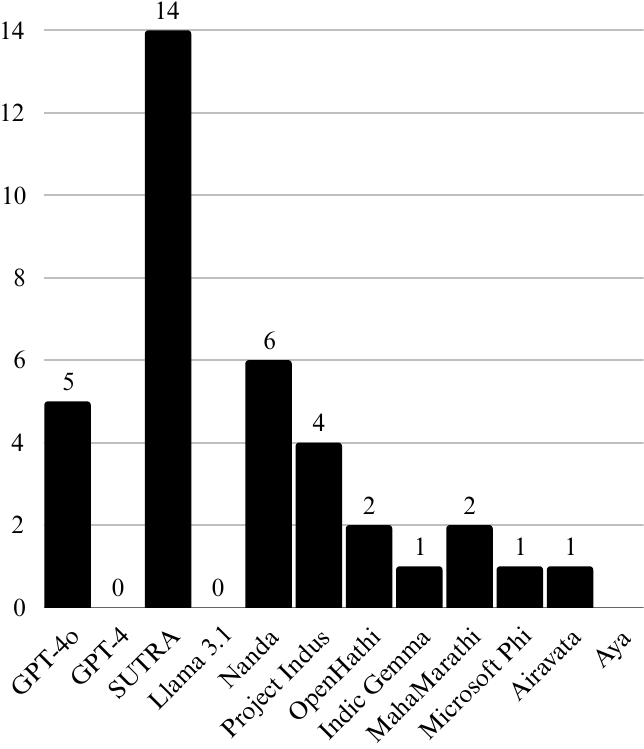}
    \caption{Number of Best Performances Achieved by Each Tokenizer Across 22 Languages.}
    \label{fig:sota-tokenizers}
\end{figure}

Figure \ref{fig:sota-tokenizers} illustrates the number of languages in which each tokenizer achieved the highest NSL score. For instance, TWO AI's SUTRA outperformed all other tokenizers in 14 languages, while MBZUAI's Nanda excelled in 6, and OpenAI's GPT-4o in 5. Other notable performers include indic models like Tech Mahindra's Project Indus with 4, Sarvam AI's OpenHathi and MahaMarathi with 2 each, and Indic Gemma, Microsoft Phi, and Airavata with 1 each.

\textbf{Number of Tokens}
Appendix \ref{sec:appendix-a} provides individual row bar charts for each language, offering a detailed breakdown of the number of tokens generated in each language. Lower token counts indicate better outcomes.

\begin{table}[htbp]
\centering
\resizebox{\textwidth}{!}{%
\begin{tabular}{@{}lcccccccccccc@{}}
\toprule
Languages & GPT-4o & GPT-4 & SUTRA & Llama 3.1 & Nanda & Project Indus & OpenHathi & Indic Gemma & MahaMarathi & Microsoft Phi & Airavata & Aya \\ \midrule
Assamese  & 0.5429  & 1.4    & \textbf{0.4571} & 1.4    & 1.4    & 2.7714  & 1.5714  & 0.8286  & 1.3143 & 1.5428 & 1.5714 & 1.5142 \\
Bengali   & 0.25    & 1.2307 & \textbf{0.2115} & 1.25   & 1.25   & 2.8076  & 1.3461  & 0.5769  & 1.0961 & 1.3269 & 1.3461 & 1.2307 \\
Bodo      & 0.5675  & 1.0540 & 0.5405          & 0.5945 & \textbf{0.4594} & 0.4864  & 0.5405  & 0.5675  & 0.5405 & 1.2432 & 0.5405 & 0.7567 \\
Dogri     & 0.5     & 1.0313 & 0.4688          & 0.5938 & \textbf{0.3750} & 0.4688  & 0.4063  & 0.4688  & 0.4063 & 1.0312 & 0.4062 & 0.7812 \\
Gujarati  & 0.47545 & 1.6875 & \textbf{0.4688} & 1.7188 & 1.7188 & 2.75    & 2.6875  & 0.7188  & 1.0938  & 2.6562 & 2.6875 & 1.7187 \\
Hindi     & 0.4091  & 1.0    & 0.4545          & 0.5909 & 0.3636 & \textbf{0.3182} & 0.4545  & 0.5455  & 0.3636 & 1.0454 & 0.4545 & 1.0454 \\
Kannada   & \textbf{0.44}    & 1.76   & \textbf{0.44}             & 1.8    & 1.8    & 2.84    & 2.52    & 0.56    & 1.12 & 2.48 & 2.52 & 1.72\\
Kashmiri  & 0.6047  & 1.093  & \textbf{0.5814} & 0.8837 & 0.8837 & 1.8605  & 1.1628  & 0.5814  & 1.186 & 1.1395 & 1.1627 & 0.9069 \\
Konkani   & \textbf{0.4643} & 1.1429 & 0.5357          & 0.6429 & 0.5     & 0.4643  & 0.6071  & 0.5357  & 1.1071 & 0.6071 & 0.6071 & 0.8214 \\
Maithili  & 0.4211  & 1.0    & 0.6316          & 0.6316 & \textbf{0.3684} & \textbf{0.3684}  & 0.5789  & 0.5789  & 1.1578 & 0.5789 & 0.5789 & 0.7368 \\
Malayalam & \textbf{0.5}    & 1.75 & \textbf{0.5} & 1.8333 & 1.8333 & 3.0     & 1.3333  & 0.6667  & 1.25 & 1.3333 & 1.3333 & 1.8333 \\
Manipuri  & 0.6471  & 1.2941 & \textbf{0.5882} & 1.3529 & 1.3529 & 2.8824  & 1.5882  & 0.7647  & 1.5882 & 1.5294 & 1.5882 & 1.2941 \\
Marathi   & 0.4706  & 0.9412 & 0.5294          & 0.6471 & \textbf{0.3529} & \textbf{0.3529}  & 0.4706  & 0.5882  & 1.0582 & 0.4706 & 0.4705 & 0.8235 \\
Nepali    & 0.4091  & 0.9091 & \textbf{0.3182} & 0.6364 & \textbf{0.3182} & 0.3636  & \textbf{0.3182}  & 0.4545  & 0.4090 & 1.1363 & \textbf{0.3182} & 0.8181 \\
Odia      & 1.0     & 2.625  & \textbf{0.625}  & 2.625  & 2.625  & 2.875   & 2.875   & 1.0625  & 2.875 & 2.8125 & 2.875 & 2.1875 \\
Punjabi   & 0.6538  & 1.6923 & \textbf{0.4615} & 1.7308 & 1.7308 & 2.7692  & 2.3077  & 0.7692  & 2.3077 & 2.2692 & 2.3076 & 1.7307 \\
Sanskrit  & \textbf{0.5}    & 1.0833 & 0.6667          & 0.5833 & \textbf{0.5}     & \textbf{0.5}     & \textbf{0.5}     & 0.6667  & 1.08333 & \textbf{0.5} & \textbf{0.5} & 0.75 \\
Santali   & 2.7647  & 2.647  & \textbf{0.4705} & 2.7058 & 2.7058 & 2.8823  & 2.8823  & 1.0588  & 2.9411 & 2.8235 & 2.8823 & 2.7058 \\
Sindhi    & \textbf{0.4117} & 0.9117 & 0.5             & 0.6176 & 0.6176 & 1.8529  & 1.0882  & 0.5588  & 1.1176 & 1.0588 & 1.0882 & 0.5882 \\
Tamil     & 0.4411  & 1.3823 & \textbf{0.3823} & 1.4117 & 1.4117 & 2.7647  & 1.2352  & 0.5294  & 1.0882  & 1.2058 & 1.2352 & 1.2058 \\
Telugu    & 0.375   & 1.75   & \textbf{0.2916} & 1.7916 & 1.7916 & 2.8333  & 2.6666  & 0.625   & 1.125 & 2.625 & 2.6666 & 1.7083 \\
Urdu      & 0.3928  & 0.7857 & \textbf{0.3571} & 0.5357 & 1.8928 & 1.8928  & 1.1071  & 0.4285  & 1.1071 & 1.0714 & 1.1071 & 0.5357 \\ \bottomrule
\end{tabular}%
}
\captionsetup{aboveskip=5pt}
\caption{
Average NSL Values Across Models for 22 Languages (lower is better). 
The bold values indicate the best-performing tokenizer for each language.
}
\label{tab:avg-nsl-values}
\end{table}
\section{Discussion}
\label{sec:discussion}
In this study, we evaluated the tokenizers from 12 LLMs in all 22 official languages of India and we found that the SUTRA tokenizer performed the best among all others, outperforming the 2nd best tokenizer by a large margin. This showcases the multilingual strength of the SUTRA tokenizer to handle the Indic languages. 

\paragraph{\textbf{Microsoft's Phi-3.5-MoE-instruct and Google's Indic Gemma}} Though both the models were developed for Indic languages, they did not perform up to the level securing best performances only in one language out of 22 languages. 

\paragraph{\textbf{Observation between GPT-4 and GPT-4o}} Another interesting observation is that GPT-4, the predecessor of GPT-4o, did not manage to secure the best tokenizer value in any of the 22 languages, a stark contrast to GPT-4o. Perhaps, this highlights that an important difference between the two models is that the newer GPT-4o is well adept at Indian languages, increasing the multi-lingual capability. 

\paragraph{\textbf{Observation of Tech Mahindra's Project Indus}} Both SUTRA and GPT-4o tokenizers manage to get a consistently low average NSL value (below 1.0) for all the languages but the Project Indus' tokenizer seems to be getting the same for only a few languages like (1) Bodo, (2) Dogri, (3) Hindi, (4) Konkani, (5) Maithili, (6) Marathi, (7) Nepali, and (8) Sanskrit. This is probably because all these 8 languages follow the same Devanagari script of writing, which the model's tokenizer was probably trained on. But for the rest of the languages, the tokenizer seems to be struggling, getting an average NSL score of above 1 (the higher the worse). 

\paragraph{\textbf{Number of Tokens}} According to the results and appendix \ref{sec:appendix-a}, tokenizers like SUTRA generate fewer tokens across the 22 Indian languages compared to others. Lower token counts suggest that the tokenizer is more efficient in processing the input text without excessive fragmentation. This is a crucial factor in improving the overall performance and computational efficiency of LLMs, particularly for large-scale applications.

\paragraph{\textbf{Significance of tokenization in LLMs}} Tokenization plays a vital role in LLMs by breaking down text into smaller units (tokens) that the model can process efficiently. A well-designed tokenizer enables the model to handle complex language structures, out-of-vocabulary words, and multi-language contexts effectively. It enhances the model's ability to understand and generate language with greater accuracy. Additionally, a good tokenizer leads to reduced computational costs and resource requirements by optimizing token generation. This results in faster training times, lower resource consumption, and overall improved performance, allowing the model to process diverse languages more effectively.

\paragraph{\textbf{Real-World Applications and Future Directions}} The insights from this study have important implications for the development of multilingual models across Indian languages. Future research could focus on enhancing tokenizers to better handle languages with complex scripts or languages with a high degree of dialectical variation, improving model performance for both high-resource and low-resource languages.
\section{Acknowledgement}
\label{sec:ack}
We would like to thank \textit{the Assam Kaziranga University} for assisting us in conducting this research.

\appendix
\section{Appendix}
\subsection{Bar Charts of Token Counts for Each Language}
\label{sec:appendix-a}



\begin{figure}[h!]
    \centering
    \includegraphics[width=\textwidth]{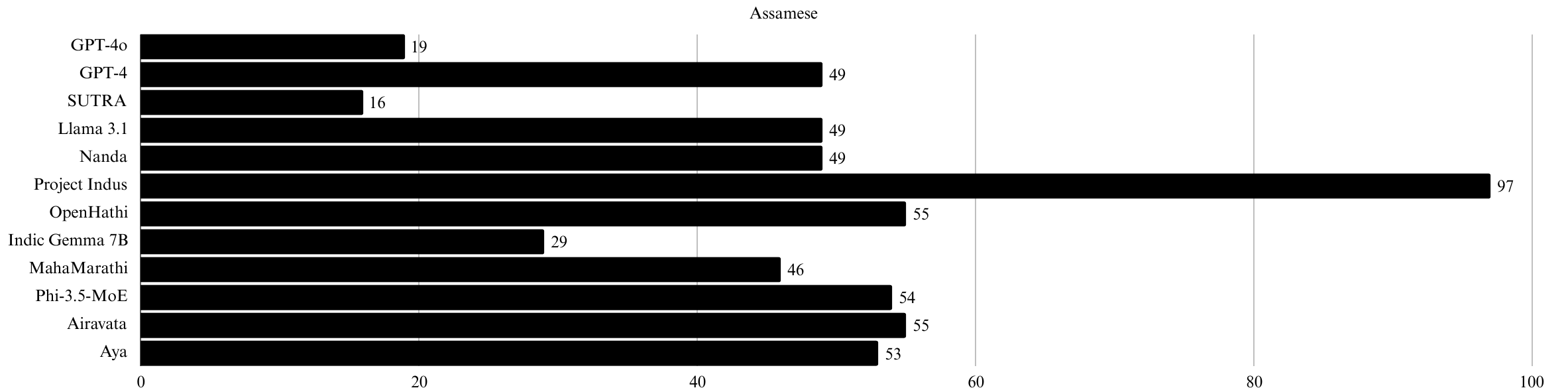}
    \caption{Number of tokens required for a single example text in Assamese. Lower values are better.}
    \label{fig:assamese-chart}
\end{figure}

\begin{figure}[h!]
    \centering
    \includegraphics[width=\textwidth]{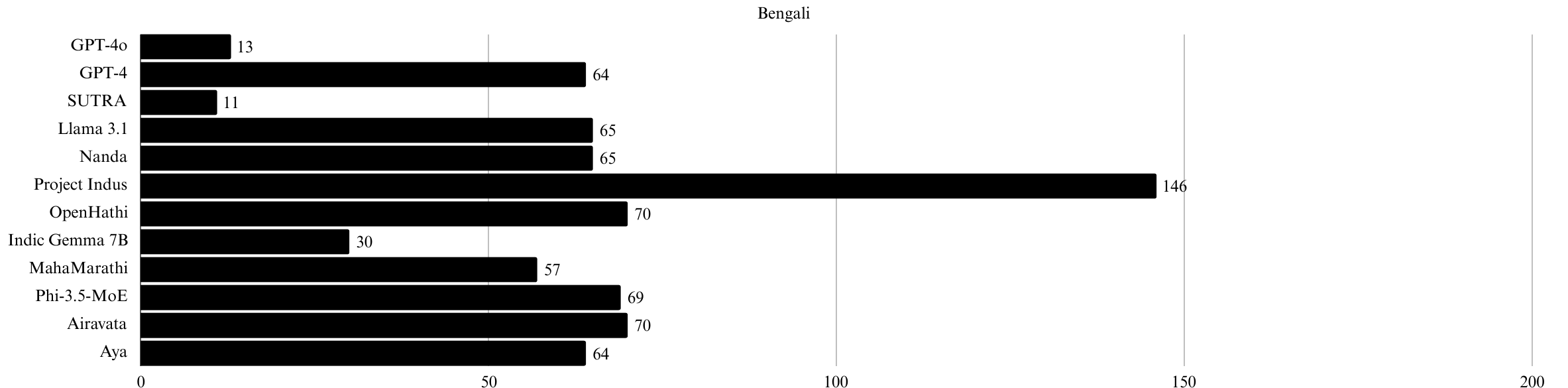}
    \caption{Number of tokens required for a single example text in Bengali. Lower values are better.}
    \label{fig:bengali-chart}
\end{figure}

\begin{figure}[h!]
    \centering
    \includegraphics[width=\textwidth]{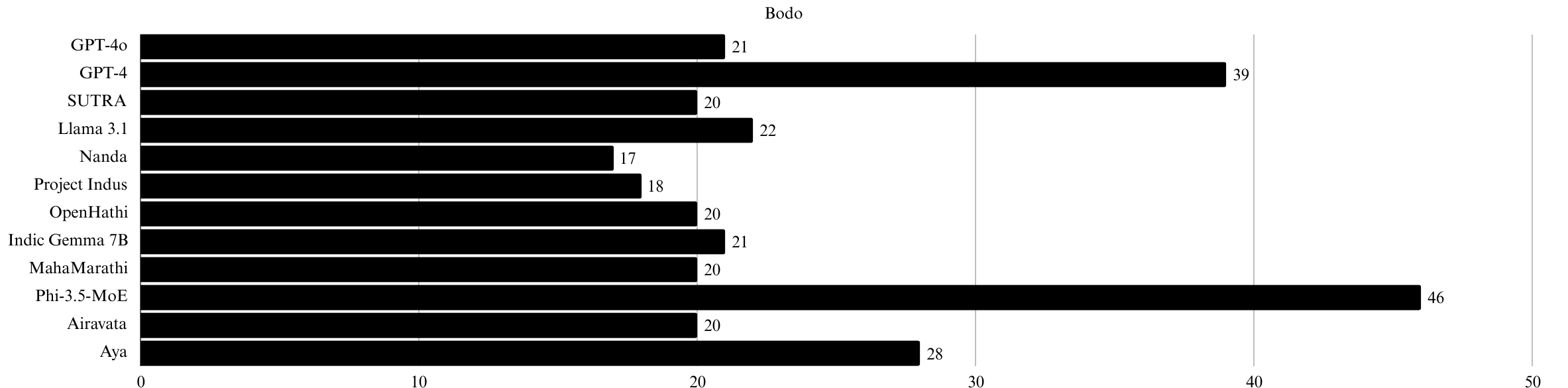}
    \caption{Number of tokens required for a single example text in Bodo. Lower values are better.}
    \label{fig:bodo-chart}
\end{figure}

\begin{figure}[h!]
    \centering
    \includegraphics[width=\textwidth]{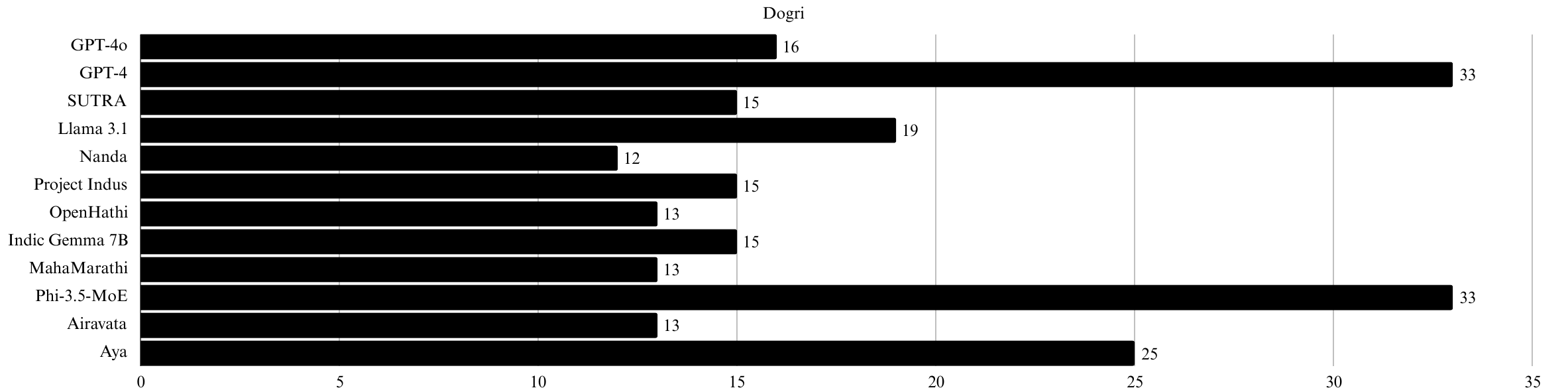}
    \caption{Number of tokens required for a single example text in Dogri. Lower values are better.}
    \label{fig:dogri-chart}
\end{figure}

\begin{figure}[h!]
    \centering
    \includegraphics[width=\textwidth]{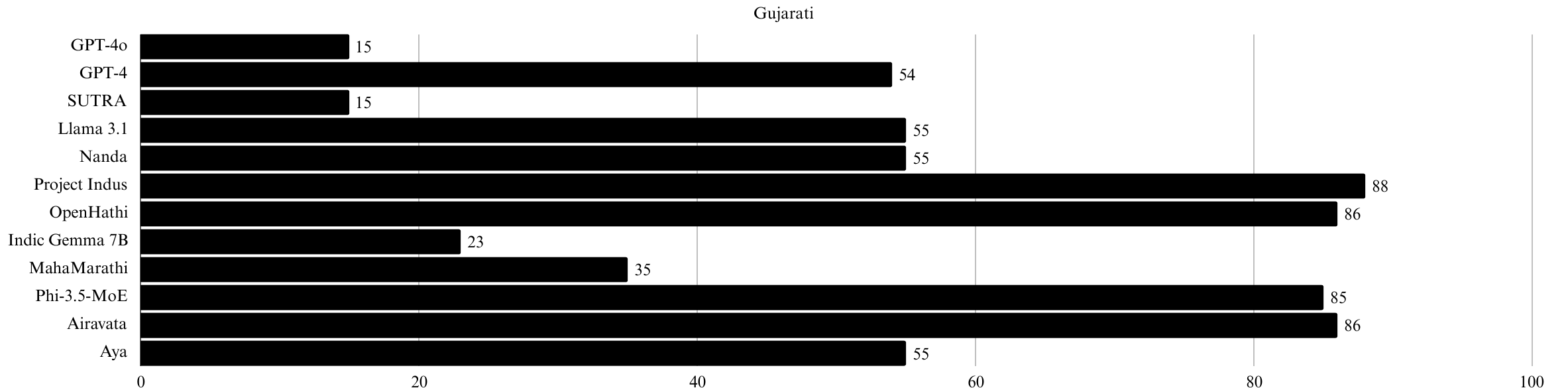}
    \caption{Number of tokens required for a single example text in Gujarati. Lower values are better.}
    \label{fig:gujarati-chart}
\end{figure}

\begin{figure}[h!]
    \centering
    \includegraphics[width=\textwidth]{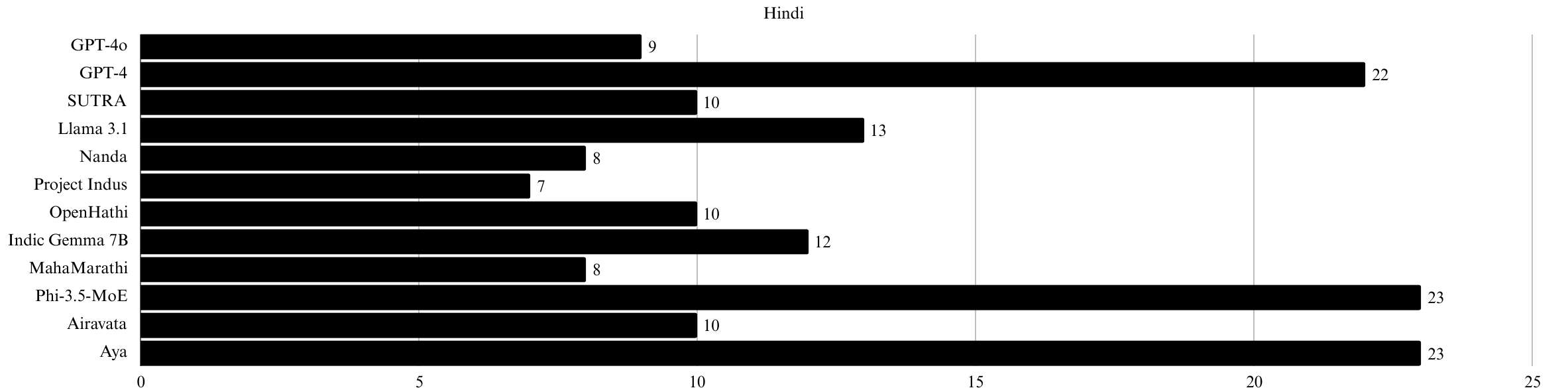}
    \caption{Number of tokens required for a single example text in Hindi. Lower values are better.}
    \label{fig:hindi-chart}
\end{figure}

\begin{figure}[h!]
    \centering
    \includegraphics[width=\textwidth]{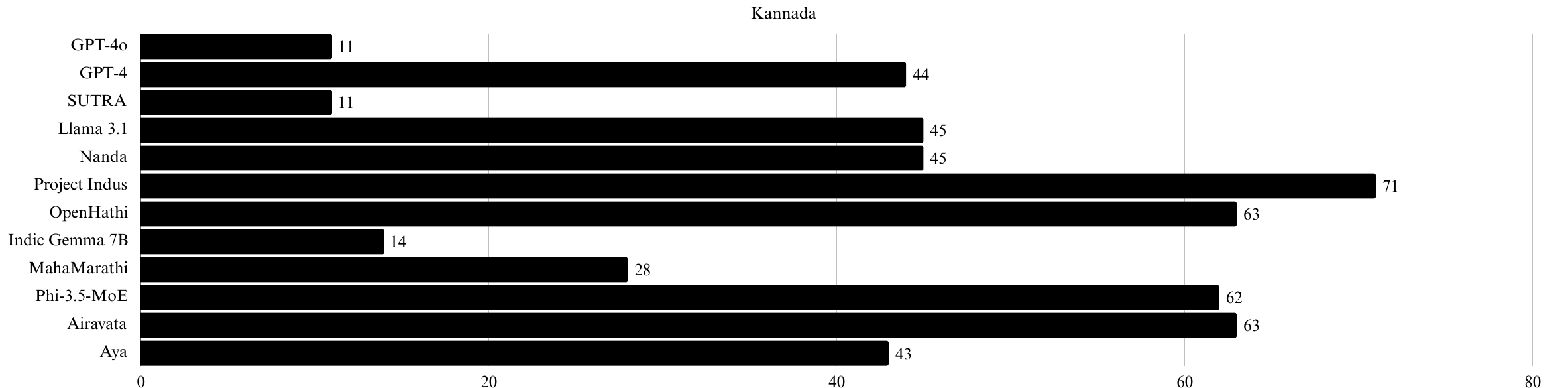}
    \caption{Number of tokens required for a single example text in Kannada. Lower values are better.}
    \label{fig:kannada-chart}
\end{figure}

\begin{figure}[h!]
    \centering
    \includegraphics[width=\textwidth]{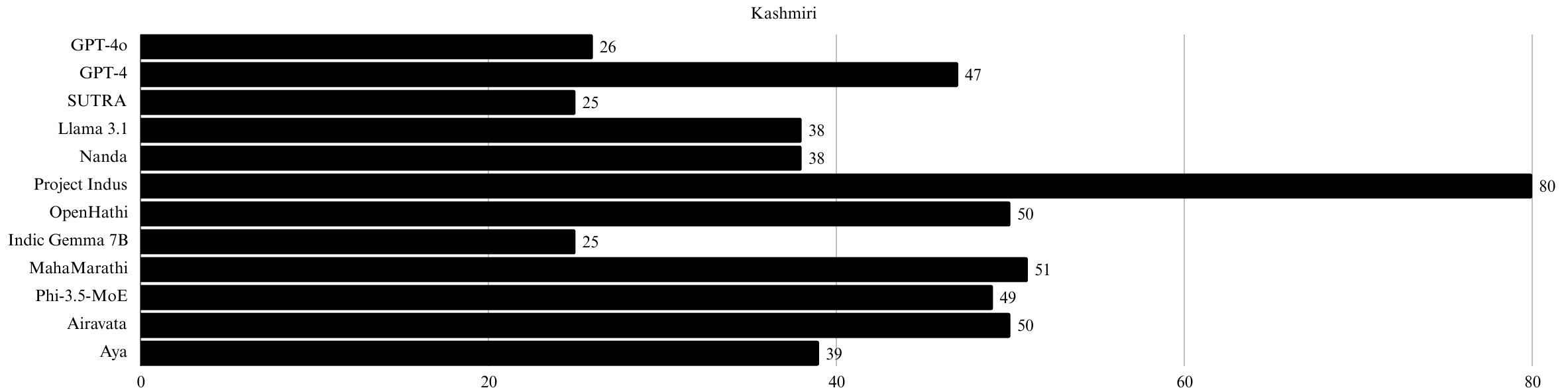}
    \caption{Number of tokens required for a single example text in Kashmiri. Lower values are better.}
    \label{fig:kashmiri-chart}
\end{figure}

\begin{figure}[h!]
    \centering
    \includegraphics[width=\textwidth]{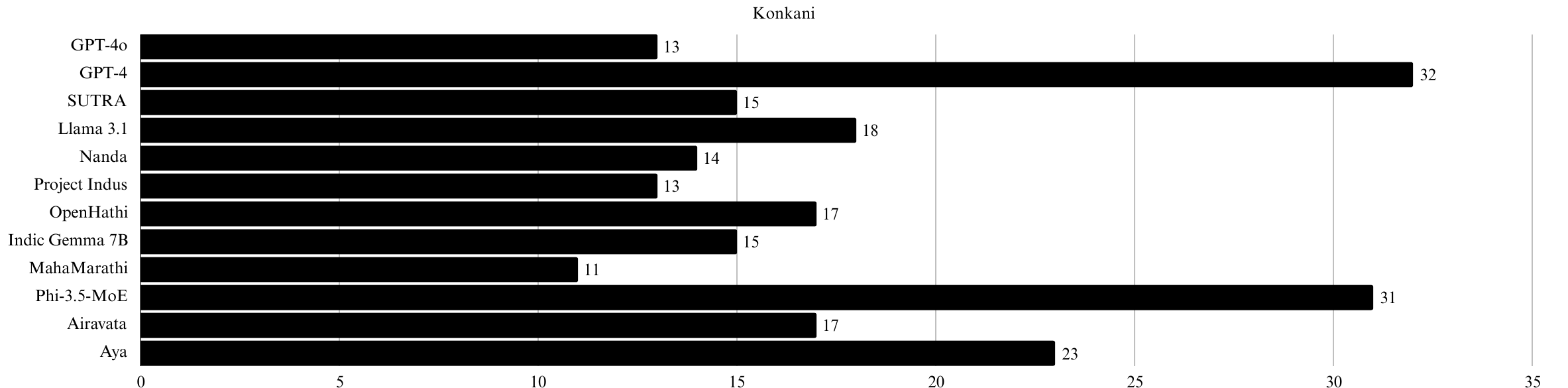}
    \caption{Number of tokens required for a single example text in Konkani. Lower values are better.}
    \label{fig:konkani-chart}
\end{figure}

\begin{figure}[h!]
    \centering
    \includegraphics[width=\textwidth]{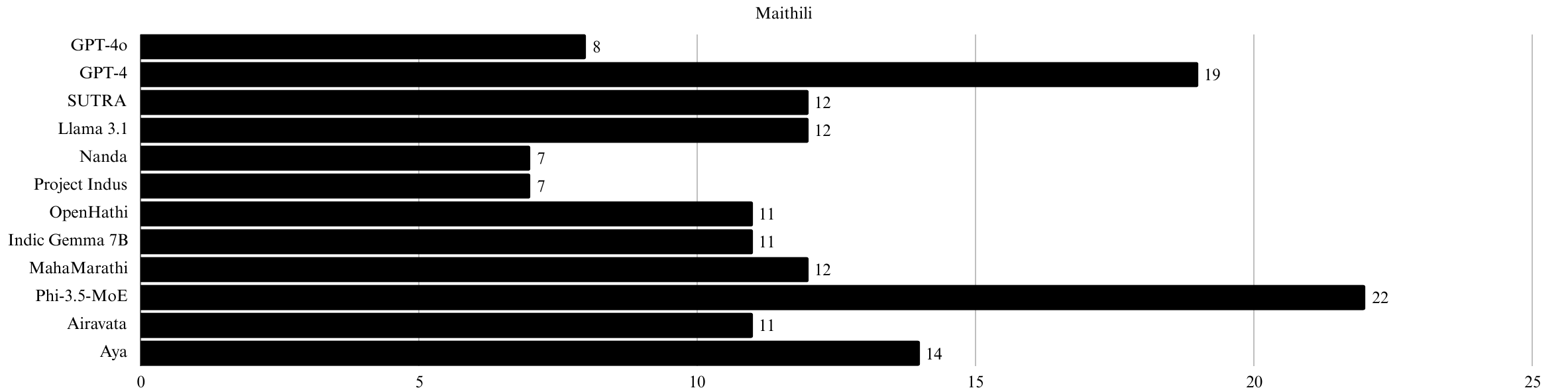}
    \caption{Number of tokens required for a single example text in Maithili. Lower values are better.}
    \label{fig:maithili-chart}
\end{figure}

\begin{figure}[h!]
    \centering
    \includegraphics[width=\textwidth]{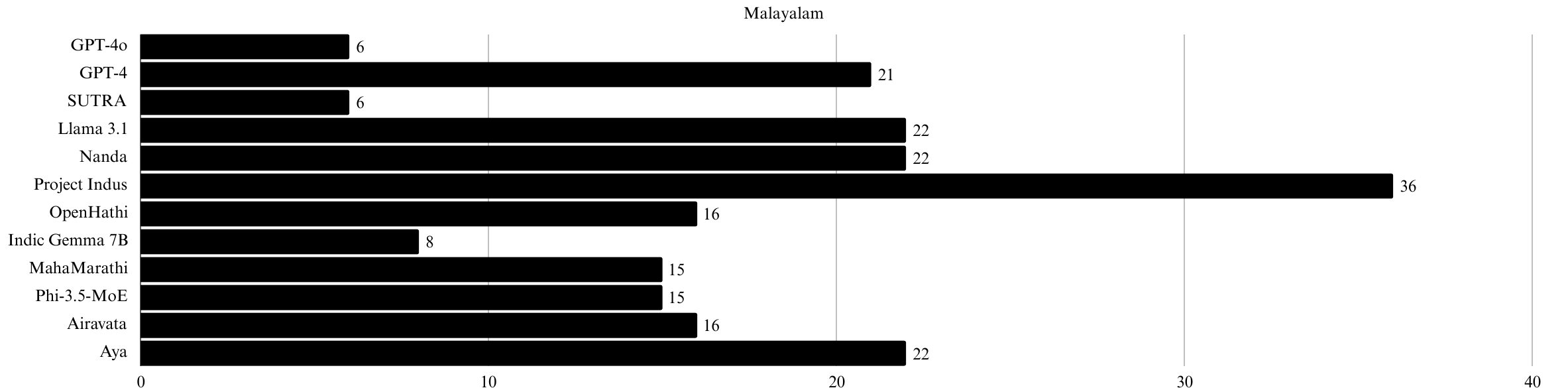}
    \caption{Number of tokens required for a single example text in Malayalam. Lower values are better.}
    \label{fig:malayalam-chart}
\end{figure}

\begin{figure}[h!]
    \centering
    \includegraphics[width=\textwidth]{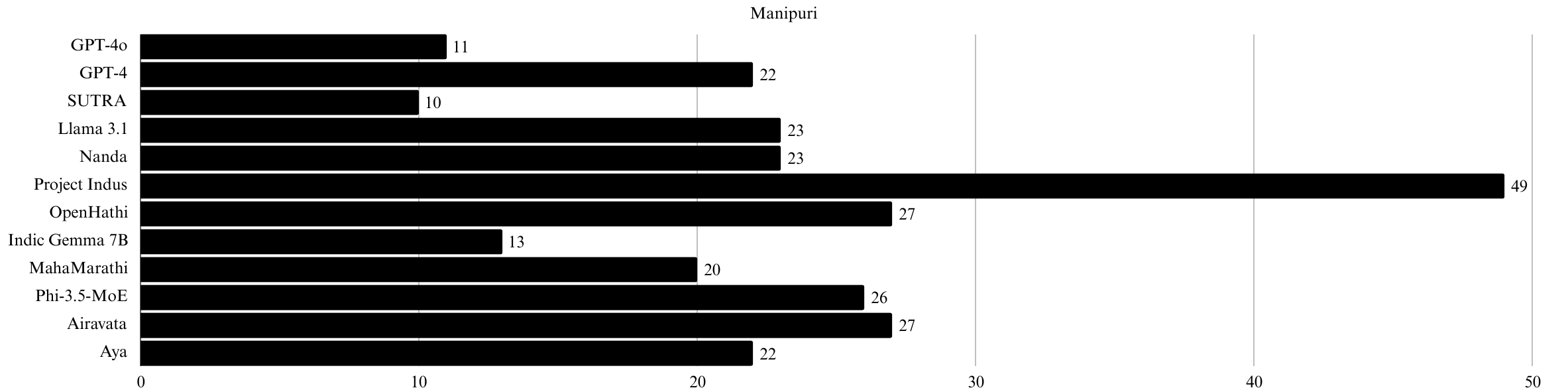}
    \caption{Number of tokens required for a single example text in Manipuri. Lower values are better.}
    \label{fig:manipuri-chart}
\end{figure}

\begin{figure}[h!]
    \centering
    \includegraphics[width=\textwidth]{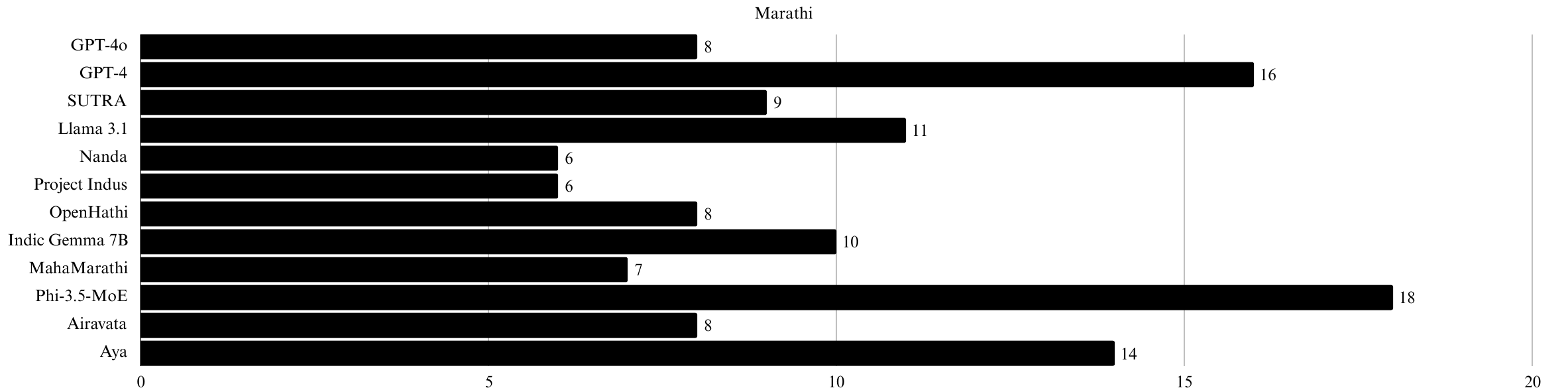}
    \caption{Number of tokens required for a single example text in Marathi. Lower values are better.}
    \label{fig:marathi-chart}
\end{figure}

\begin{figure}[h!]
    \centering
    \includegraphics[width=\textwidth]{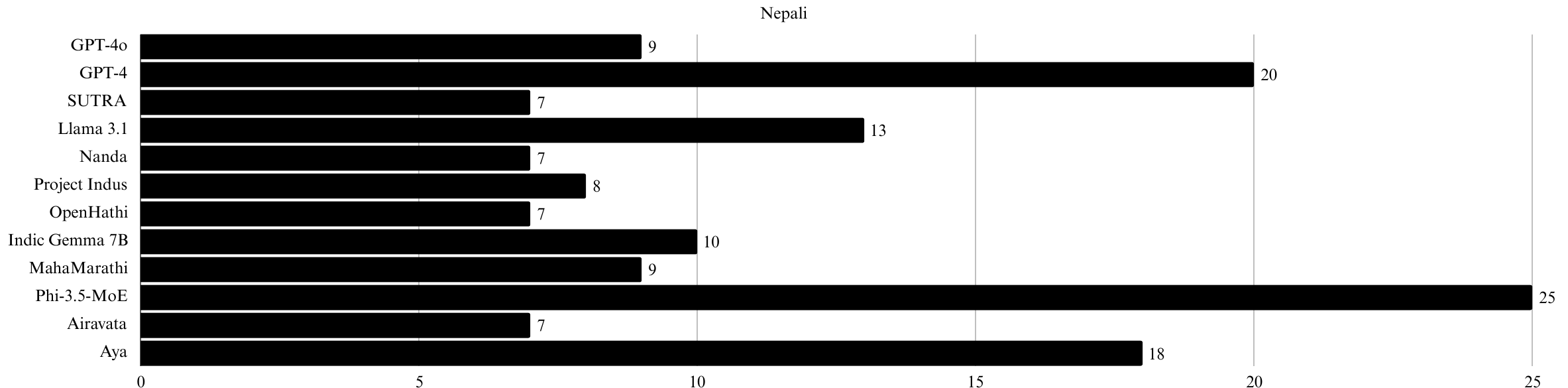}
    \caption{Number of tokens required for a single example text in Nepali. Lower values are better.}
    \label{fig:nepali-chart}
\end{figure}

\begin{figure}[h!]
    \centering
    \includegraphics[width=\textwidth]{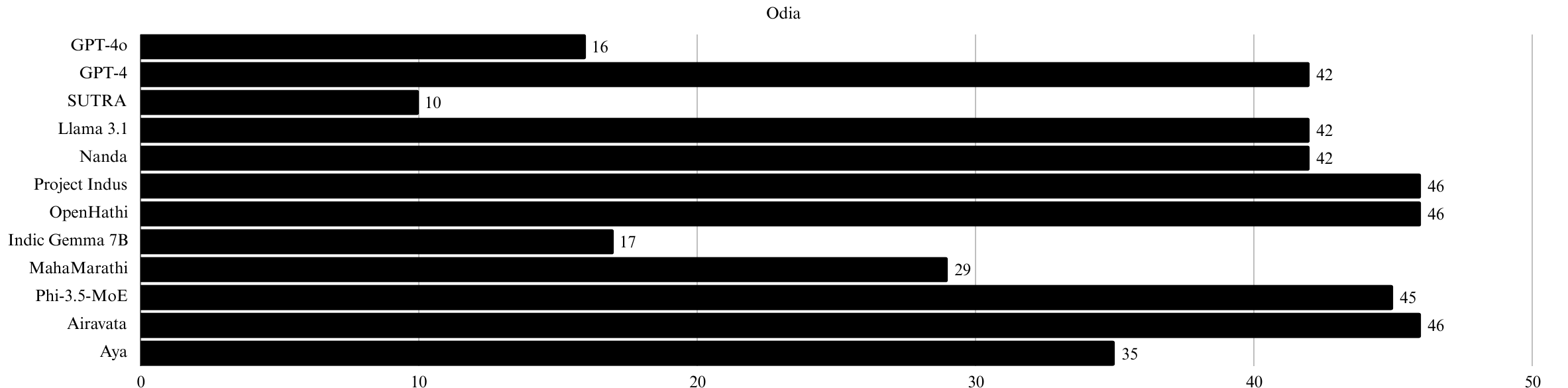}
    \caption{Number of tokens required for a single example text in Odia. Lower values are better.}
    \label{fig:odia-chart}
\end{figure}

\begin{figure}[h!]
    \centering
    \includegraphics[width=\textwidth]{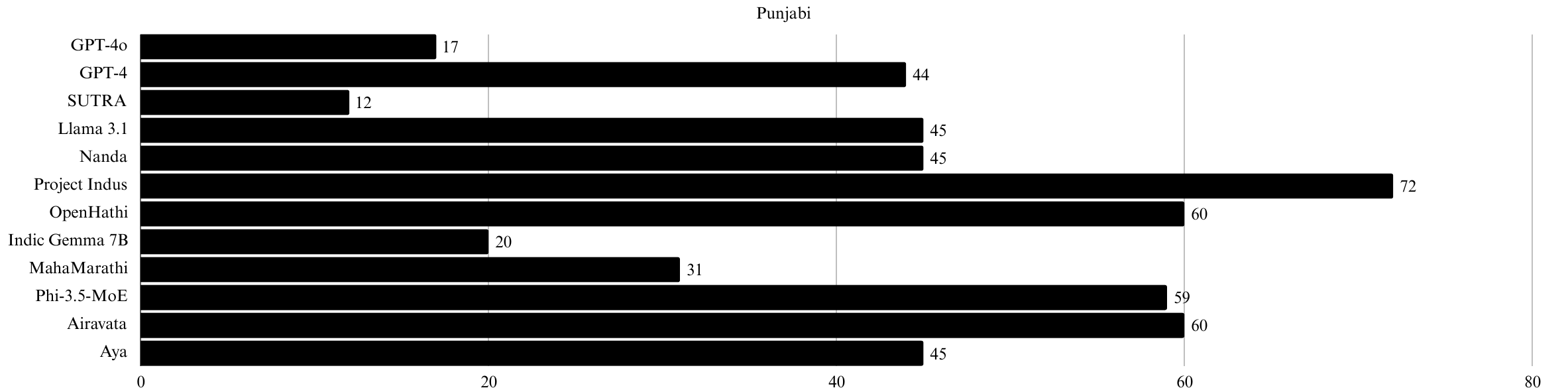}
    \caption{Number of tokens required for a single example text in Punjabi. Lower values are better.}
    \label{fig:punjabi-chart}
\end{figure}

\begin{figure}[h!]
    \centering
    \includegraphics[width=\textwidth]{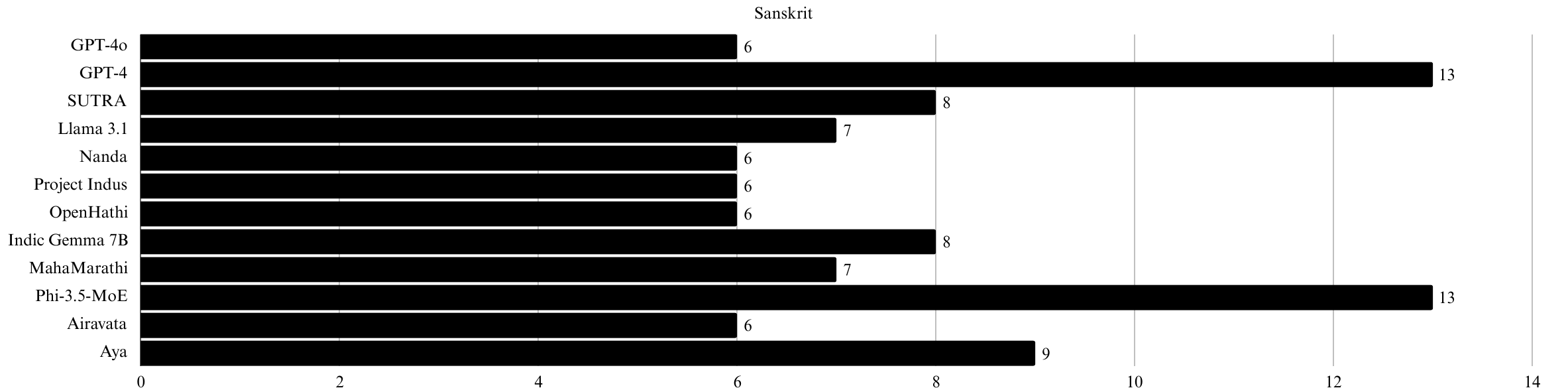}
    \caption{Number of tokens required for a single example text in Sanskrit. Lower values are better.}
    \label{fig:sanskrit-chart}
\end{figure}

\begin{figure}[h!]
    \centering
    \includegraphics[width=\textwidth]{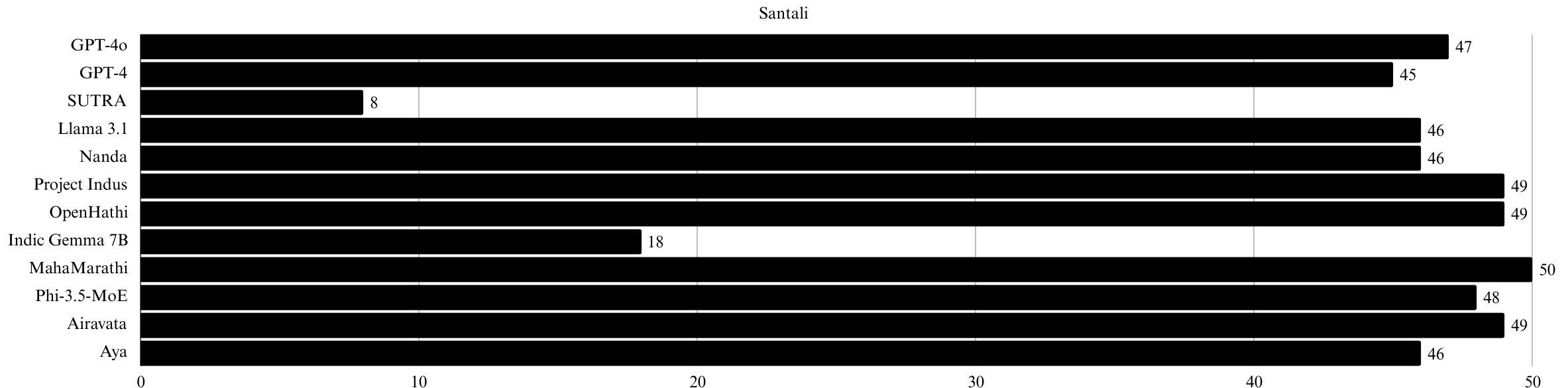}
    \caption{Number of tokens required for a single example text in Santali. Lower values are better.}
    \label{fig:santali-chart}
\end{figure}

\begin{figure}[h!]
    \centering
    \includegraphics[width=\textwidth]{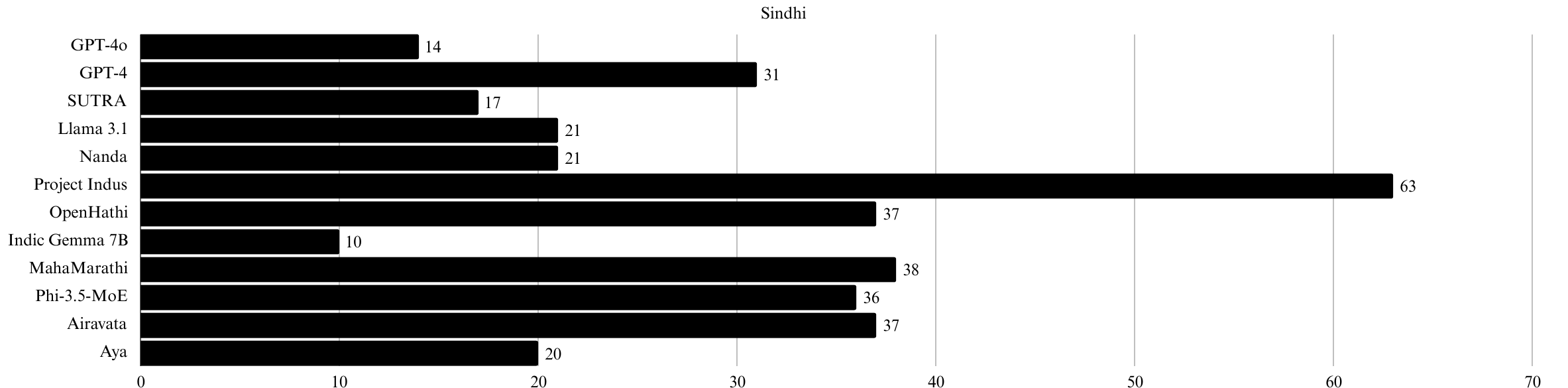}
    \caption{Number of tokens required for a single example text in Sindhi. Lower values are better.}
    \label{fig:sindhi-chart}
\end{figure}

\begin{figure}[h!]
    \centering
    \includegraphics[width=\textwidth]{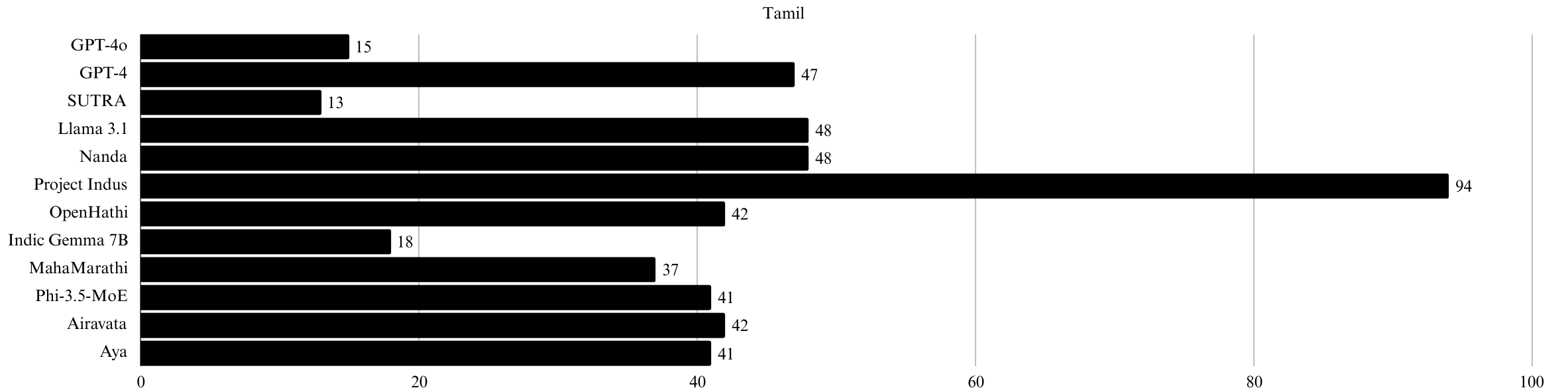}
    \caption{Number of tokens required for a single example text in Tamil. Lower values are better.}
    \label{fig:tamil-chart}
\end{figure}

\begin{figure}[h!]
    \centering
    \includegraphics[width=\textwidth]{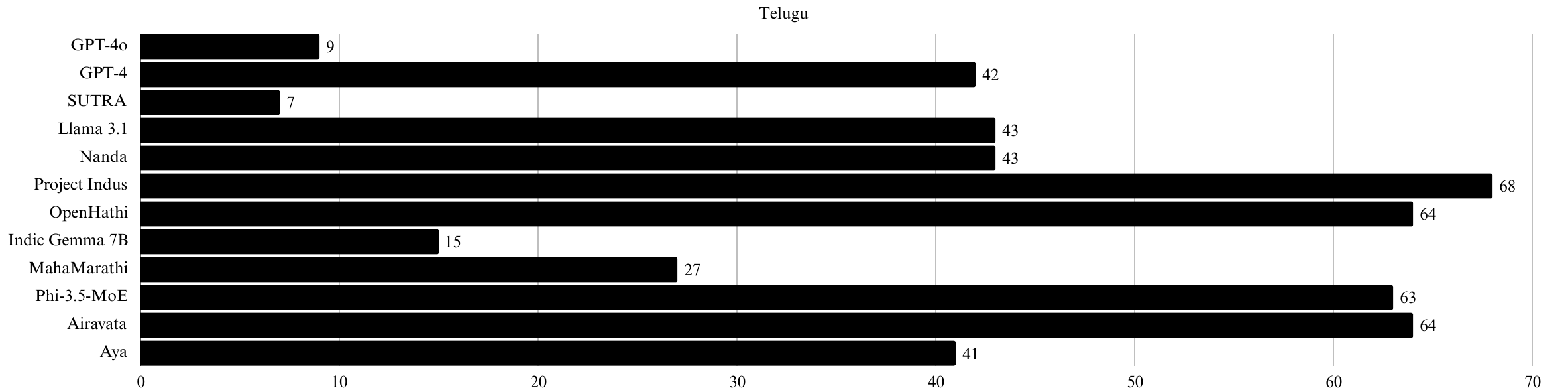}
    \caption{Number of tokens required for a single example text in Telugu. Lower values are better.}
    \label{fig:telugu-chart}
\end{figure}

\begin{figure}[h!]
    \centering
    \includegraphics[width=\textwidth]{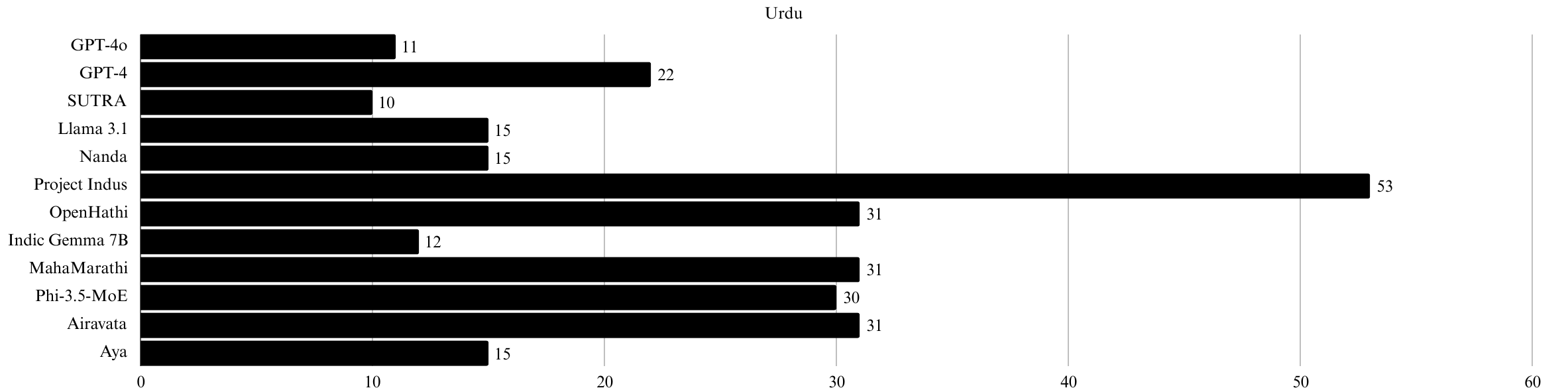}
    \caption{Number of tokens required for a single example text in Urdu. Lower values are better.}
    \label{fig:urdu-chart}
\end{figure}
\subsection{Example Texts Used for Tokenizer Evaluation}
\label{sec:appendix-b}
\begin{figure}[h!]
    \centering
    \includegraphics[width=\textwidth]{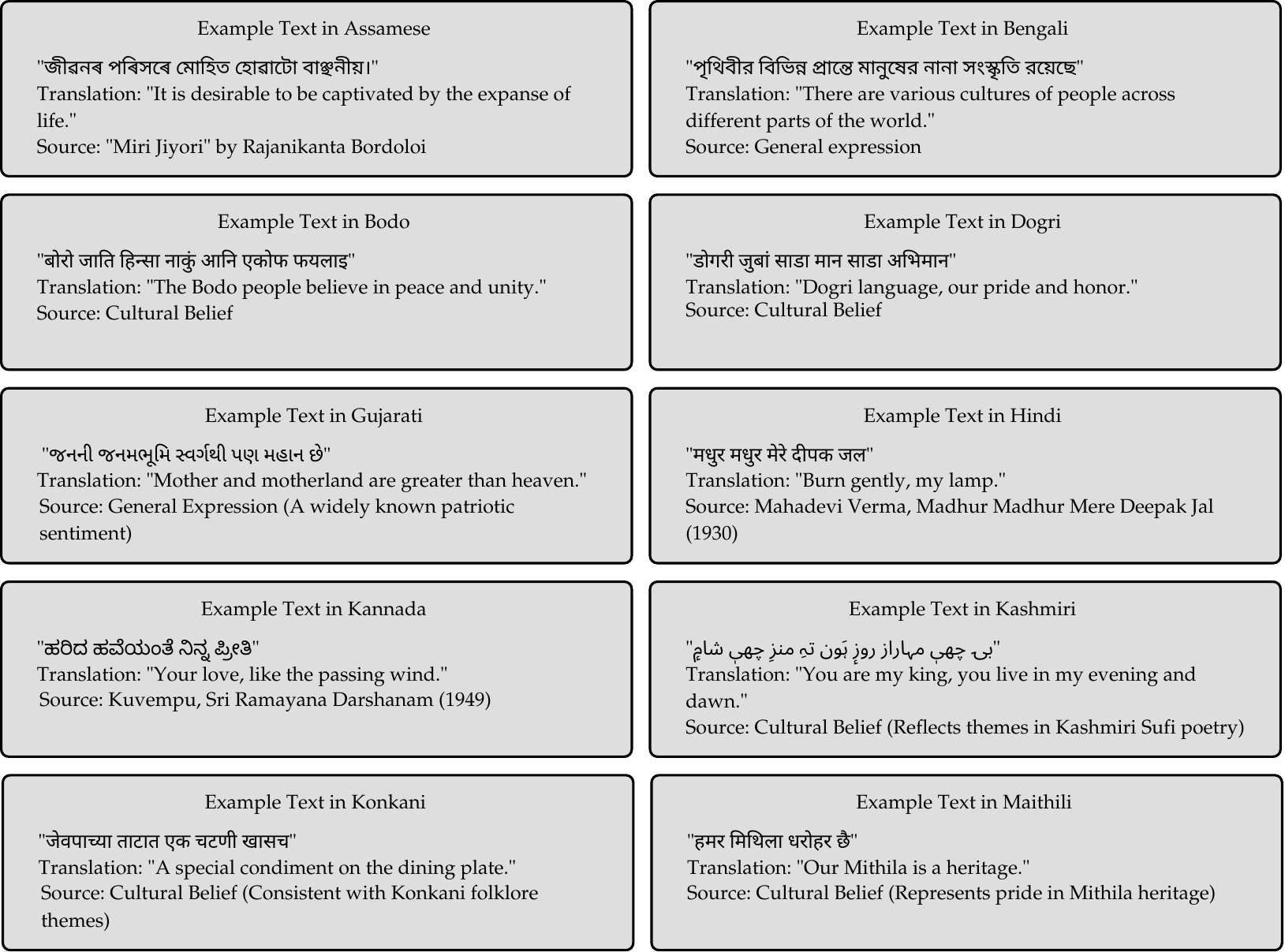}
    \caption{Example Texts for Assamese, Bengali, Bodo, Dogri, Gujarati, Hindi, Kannada, Kashmiri, Konkani, Maithili.}
    \label{fig:example-texts-1st-half}
\end{figure}

\begin{figure}[h!]
    \centering
    \includegraphics[width=\textwidth]{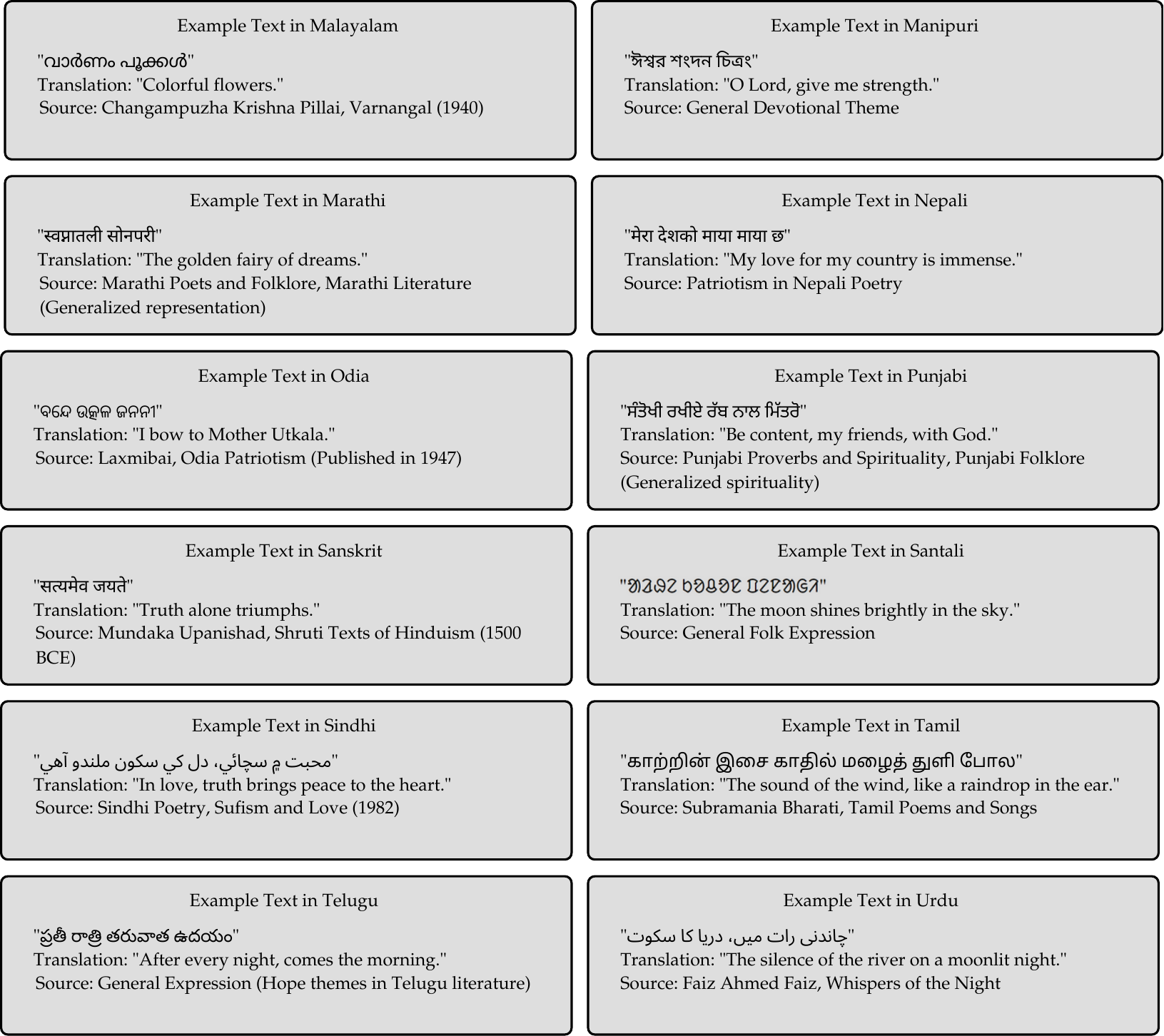}
    \caption{Example Texts for Maithili, Malayalam, Manipuri, Marathi, Nepali, Odia, Punjabi, Sanskrit, Santali, Sindhi, Tamil, Telugu, Urdu.}
    \label{fig:example-texts-2nd-half}
\end{figure}


\begin{thebibliography}{99}

\bibitem{CHIARELLO2024103002}
F. Chiarello, V. Giordano, I. Spada, S. Barandoni, and G. Fantoni, "Future applications of generative large language models: A data-driven case study on ChatGPT," \textit{Technovation}, vol. 133, p. 103002, 2024. [Online]. Available: \url{https://www.sciencedirect.com/science/article/pii/S016649722400052X}. [Accessed: Nov. 12, 2024].

\bibitem{nie2024surveylargelanguagemodels}
Y. Nie, Y. Kong, X. Dong, J. M. Mulvey, H. V. Poor, Q. Wen, and S. Zohren, "A Survey of Large Language Models for Financial Applications: Progress, Prospects and Challenges," \textit{arXiv preprint arXiv:2406.11903}, 2024. [Online]. Available: \url{https://arxiv.org/abs/2406.11903}. [Accessed: Nov. 12, 2024].

\bibitem{tamang2024performanceevaluationtokenizerslarge}
S. Tamang and D. J. Bora, "Performance Evaluation of Tokenizers in Large Language Models for the Assamese Language," arXiv preprint arXiv:2410.03718, 2024. [Online]. Available: \url{https://arxiv.org/abs/2410.03718}

\bibitem{yang2024largelanguagemodeltokenizer}
J. Yang, Z. Wang, Y. Lin, and Z. Zhao, "Large Language Model Tokenizer Bias: A Case Study and Solution on GPT-4o," \textit{arXiv preprint arXiv:2406.11214}, 2024. [Online]. Available: \url{https://arxiv.org/abs/2406.11214}. [Accessed: Nov. 13, 2024].

\bibitem{song2021fastwordpiecetokenization}
X. Song, A. Salcianu, Y. Song, D. Dopson, and D. Zhou, “Fast WordPiece Tokenization,” 2021. [Online]. Available: \url{https://arxiv.org/abs/2012.15524}.

\bibitem{ogundepo2022betterwhitespaceinformationretrieval}
O. Ogundepo, X. Zhang, and J. Lin, “Better Than Whitespace: Information Retrieval for Languages without Custom Tokenizers,” 2022. [Online]. Available: \url{https://arxiv.org/abs/2210.05481}.

\bibitem{kozma2024theoreticalanalysisbytepairencoding}
L. Kozma and J. Voderholzer, “Theoretical Analysis of Byte-Pair Encoding,” 2024. [Online]. Available: \url{https://arxiv.org/abs/2411.08671}.

\bibitem{zouhar2024formalperspectivebytepairencoding}
V. Zouhar, C. Meister, J. L. Gastaldi, L. Du, T. Vieira, M. Sachan, and R. Cotterell, “A Formal Perspective on Byte-Pair Encoding,” 2024. [Online]. Available: \url{https://arxiv.org/abs/2306.16837}.

\bibitem{singh2024indicgenbenchmultilingualbenchmarkevaluate}
H. Singh, N. Gupta, S. Bharadwaj, D. Tewari, and P. Talukdar, "IndicGenBench: A Multilingual Benchmark to Evaluate Generation Capabilities of LLMs on Indic Languages," \textit{arXiv preprint arXiv:2404.16816}, 2024. [Online]. Available: \url{https://arxiv.org/abs/2404.16816}.

\bibitem{euTokenizerPerformance}
Occiglot, "EU Tokenizer Performance," [Online]. Available: \url{https://occiglot.eu/posts/eu_tokenizer_perfomance/}. Accessed: Nov. 17, 2024.

\bibitem{eu_tokenizer_perfomance}
Occiglot, "Tokenizer performance on EU languages," Occiglot Blog, Sep. 26, 2023. [Online]. Available: \url{https://occiglot.eu/posts/eu_tokenizer_perfomance/}.

\bibitem{dagan2024gettingtokenizerpretrainingdomain}
G. Dagan, G. Synnaeve, and B. Rozière, "Getting the most out of your tokenizer for pre-training and domain adaptation," \textit{arXiv preprint arXiv:2402.01035}, 2024. [Online]. Available: \url{https://arxiv.org/abs/2402.01035}.

\bibitem{microsoft2024gpt4o}
Microsoft, "Exploring the New Frontier of AI: OpenAI's GPT-4-O for Indic Languages," Azure AI Blog, Oct. 30, 2024. [Online]. Available: \url{https://techcommunity.microsoft.com/blog/azure-ai-services-blog/exploring-the-new-frontier-of-ai-openais-gpt-4-o-for-indic-languages/4142383}.

\bibitem{openai2024gpt4technicalreport}
OpenAI, J. Achiam, S. Adler, S. Agarwal, L. Ahmad, \textit{et al.}, "GPT-4 Technical Report," arXiv preprint arXiv:2303.08774, 2024. [Online]. Available: \url{https://arxiv.org/abs/2303.08774}.

\bibitem{bendale2024sutrascalablemultilinguallanguage}
A. Bendale, M. Sapienza, S. Ripplinger, S. Gibbs, J. Lee, and P. Mistry, "SUTRA: Scalable Multilingual Language Model Architecture," arXiv preprint arXiv:2405.06694, 2024. [Online]. Available: \url{https://arxiv.org/abs/2405.06694}.

\bibitem{geminiteam2024geminifamilyhighlycapable}
Gemini Team et al., 
\textit{Gemini: A Family of Highly Capable Multimodal Models}, 
2024. Available at: \url{https://arxiv.org/abs/2401.12345}.

\bibitem{dubey2024llama3herdmodels}
Dubey, Abhimanyu, et al. 
\textit{The Llama 3 Herd of Models}. 
arXiv preprint, 2024. Available at: \url{https://arxiv.org/abs/2407.21783}.

\bibitem{minaee2024largelanguagemodelssurvey}
Shervin Minaee, Tomas Mikolov, Narjes Nikzad, Meysam Chenaghlu, Richard Socher, Xavier Amatriain, and Jianfeng Gao, "Large Language Models: A Survey," arXiv preprint arXiv:2402.06196, 2024. [Online]. Available: \url{https://arxiv.org/abs/2402.06196}

\bibitem{naveed2024comprehensiveoverviewlargelanguage}
Humza Naveed, Asad Ullah Khan, Shi Qiu, Muhammad Saqib, Saeed Anwar, Muhammad Usman, Naveed Akhtar, Nick Barnes, and Ajmal Mian, "A Comprehensive Overview of Large Language Models," arXiv preprint arXiv:2307.06435, 2024. [Online]. Available: \url{https://arxiv.org/abs/2307.06435}

\bibitem{adasci2024multilingualtokenization}
AdaSci, "Multilingual Tokenization Efficiency in Large Language Models: A Study on Indian Languages," [Online]. Available: \url{https://adasci.org/multilingual-tokenization-efficiency-in-large-language-models-a-study-on-indian-languages/}

\bibitem{toraman2023impacttokenization}
Cagri Toraman, Eyup Halit Yilmaz, Furkan Şahinüç, and Oguzhan Ozcelik, 
"Impact of Tokenization on Language Models: An Analysis for Turkish," 
ACM Transactions on Asian and Low-Resource Language Information Processing (TALLIP), 
vol. 22, no. 4, article 116, pp. 1–21, Mar. 2023. 
[Online]. Available: \url{https://doi.org/10.1145/3578707}

\bibitem{bhat2023indic}
Bhat, Savita, Vasudeva Varma, and Niranjan Pedanekar. 
"Generative Models For Indic Languages: Evaluating Content Generation Capabilities." 
In *Proceedings of the 14th International Conference on Recent Advances in Natural Language Processing*, 
edited by Ruslan Mitkov and Galia Angelova, 187--195. Varna, Bulgaria: INCOMA Ltd., September 2023. 
\url{https://aclanthology.org/2023.ranlp-1.21}.

\bibitem{singh2024indicgenbench}
Singh, Harman, Nitish Gupta, Shikhar Bharadwaj, Dinesh Tewari, and Partha Talukdar. 
"IndicGenBench: A Multilingual Benchmark to Evaluate Generation Capabilities of LLMs on Indic Languages." 
In *Proceedings of the 62nd Annual Meeting of the Association for Computational Linguistics (Volume 1: Long Papers)*, 
edited by Lun-Wei Ku, Andre Martins, and Vivek Srikumar, 11047--11073. Bangkok, Thailand: Association for Computational Linguistics, August 2024. 
\url{https://aclanthology.org/2024.acl-long.595}, 
DOI: 10.18653/v1/2024.acl-long.595.


\bibitem{singh2024indicqabench}
Singh, Abhishek Kumar, Rudra Murthy, Vishwajeet Kumar, Jaydeep Sen, and Ganesh Ramakrishnan. 
"Indic QA Benchmark: A Multilingual Benchmark to Evaluate Question Answering Capability of LLMs for Indic Languages." 
arXiv preprint, 2024. 
\url{https://arxiv.org/abs/2407.13522}.


\bibitem{kumar2022indicnlg}
Kumar, Aman, Himani Shrotriya, Prachi Sahu, Amogh Mishra, Raj Dabre, Ratish Puduppully, Anoop Kunchukuttan, Mitesh M. Khapra, and Pratyush Kumar. 
"IndicNLG Benchmark: Multilingual Datasets for Diverse NLG Tasks in Indic Languages." 
In *Proceedings of the 2022 Conference on Empirical Methods in Natural Language Processing*, 
edited by Yoav Goldberg, Zornitsa Kozareva, and Yue Zhang, 5363--5394. Abu Dhabi, United Arab Emirates: Association for Computational Linguistics, December 2022. 
\url{https://aclanthology.org/2022.emnlp-main.360}, 
DOI: 10.18653/v1/2022.emnlp-main.360.

\bibitem{alyafeai2021evaluatingtokenizers}
Alyafeai, Zaid, Maged S. Al-shaibani, Mustafa Ghaleb, and Irfan Ahmad. 
"Evaluating Various Tokenizers for Arabic Text Classification." 
arXiv preprint, 2021. 
\url{https://arxiv.org/abs/2106.07540}.

\bibitem{eighthSchedule}
Government of India, "Eighth Schedule," [Online]. Available: \url{https://www.mea.gov.in/Images/pdf1/S8.pdf}. Accessed: Dec. 5, 2023.

\bibitem{watts2024parikshalargescaleinvestigationhumanllm}
I. Watts, V. Gumma, A. Yadavalli, V. Seshadri, M. Swaminathan, and S. Sitaram, 
``PARIKSHA: A Large-Scale Investigation of Human-LLM Evaluator Agreement on Multilingual and Multi-Cultural Data,'' 
\emph{arXiv preprint arXiv:2406.15053}, 2024. [Online]. Available: \url{https://arxiv.org/abs/2406.15053}

\bibitem{twoai_sutra}
DeepLearning.ai, 
\emph{Startup TWO AI Launches SUTRA: A Multilingual Model for South Asian Markets}, 
\url{https://www.deeplearning.ai/the-batch/startup-two-ai-launches-sutra-a-multilingual-model-for-south-asian-markets/}, 
Accessed: 2024-11-18.

\end{thebibliography}
\end{document}